%% file: root.tex
\documentclass[letterpaper, 10 pt, conference]{ieeeconf}  % Comment this line out if you need a4paper

\IEEEoverridecommandlockouts                              % This command is only needed if 
\usepackage{amsmath,amssymb,balance,booktabs,cite,changes,color,enumerate,float,graphicx,hyperref,multirow,siunitx,graphbox}
\usepackage[font=footnotesize]{caption}
\input{defs.tex}
\usepackage{algorithm}
\usepackage[noend]{algpseudocode}

\definecolor{blue}{rgb}{0.44, 0.65, 0.82}
\definechangesauthor[color=blue, name=laura]{LS}
\usepackage{cleveref}
\usepackage{subcaption}

\title{\LARGE \bf A Walk in the Park: Learning to Walk in 20 Minutes\\ With Model-Free Reinforcement Learning}

\author{
\authorblockN{Laura Smith\textsuperscript{* 1}, Ilya Kostrikov\textsuperscript{* 1}, Sergey Levine\textsuperscript{1}}
\authorblockA{\textsuperscript{*}Equal contribution \textsuperscript{1}Berkeley AI Research, UC Berkeley  \\
\texttt{\{smithlaura, kostrikov\}@berkeley.edu, svlevine@eecs.berkeley.edu}}
}

\begin{document}
% Reduce space after algorithms
\setlength{\textfloatsep}{7pt}

\makeatletter
\let\@oldmaketitle\@maketitle%
\renewcommand{\@maketitle}{\@oldmaketitle%
    \centering
    \includegraphics[width=.9\linewidth]{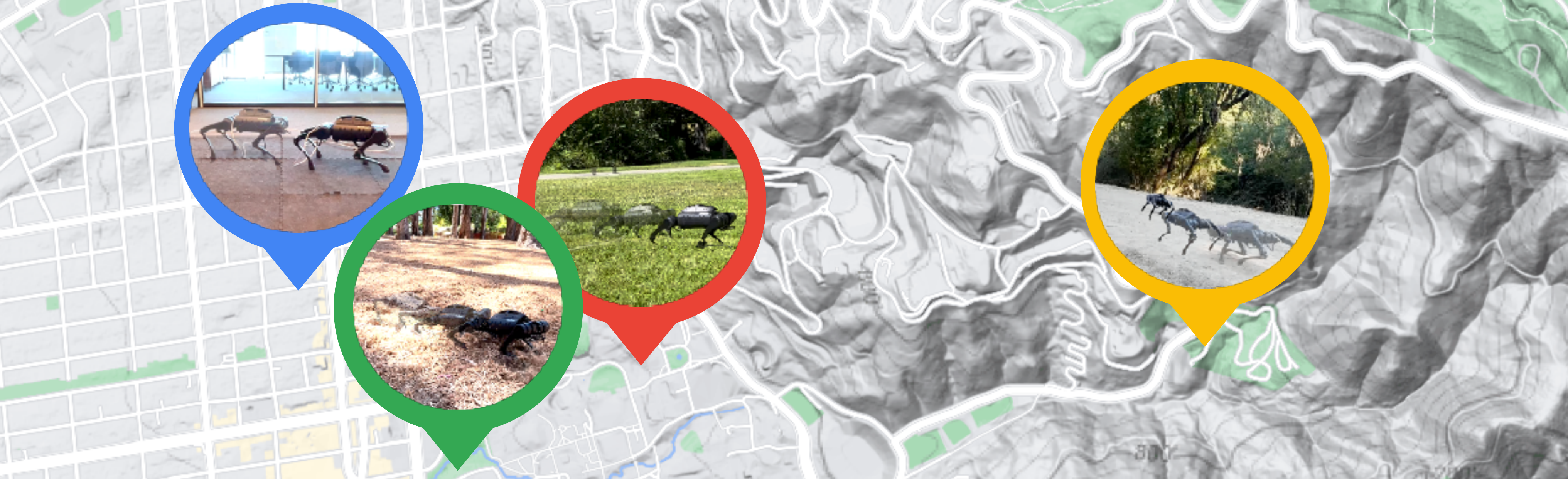}
    \captionof{figure}{\footnotesize We demonstrate that deep reinforcement learning can be used to efficiently train a quadruped robot directly on various real world terrains, e.g., flat ground (blue); soft, irregular mulch (green); grass (red); and a hiking trail (yellow), acquiring effective gaits within 20 minutes of training.}
    \vspace{-.4cm}
    \label{fig:teaser}
}
\makeatother

\maketitle
\thispagestyle{empty}
\pagestyle{empty}

%%%%%%%%%%%%%%%%%%%%%%%%%%%%%%%%%%%%%%%%%%%%%%%%%%%%%%%%%%%%%%%%%%%%%%%%%%%%%%%%
\begin{abstract} Deep reinforcement learning is a promising approach to learning policies in uncontrolled environments that do not require domain knowledge. Unfortunately, due to sample inefficiency, deep RL applications have primarily focused on simulated environments. In this work, we demonstrate that the recent advancements in machine learning algorithms and libraries combined with a carefully tuned robot controller lead to learning quadruped locomotion in only 20 minutes in the real world. We evaluate our approach on several indoor and outdoor terrains which are known to be challenging for classical model-based controllers. We observe the robot to be able to learn walking gait consistently on all of these terrains. Finally, we evaluate our design decisions in a simulated environment. 
We provide videos of all real-world training and code to reproduce our results on our website: \tt \url{https://sites.google.com/berkeley.edu/walk-in-the-park} 
\end{abstract}

%%%%%%%%%%%%%%%%%%%%%%%%%%%%%%%%%%%%%%%%%%%%%%%%%%%%%%%%%%%%%%%%%%%%%%%%%%%%%%%%

\input{sections/intro}
\input{sections/relatedwork.tex}

\input{sections/preliminaries.tex}

\input{sections/system.tex}

\input{sections/experiments.tex}

\input{sections/discussion.tex}

\input{sections/acknowledgements}

\balance
\bibliography{references}
\bibliographystyle{IEEEtran}

\end{document}

%% file: defs.tex
\newcommand{\alert}[1]{\textbf{}}

\newcommand{\BEAS}{\begin{eqnarray*}}
\newcommand{\EEAS}{\end{eqnarray*}}
\newcommand{\BEA}{\begin{eqnarray}}
\newcommand{\EEA}{\end{eqnarray}}
\newcommand{\BEQ}{\begin{equation}}
\newcommand{\EEQ}{\end{equation}}
\newcommand{\BIT}{\begin{itemize}}
\newcommand{\EIT}{\end{itemize}}
\newcommand{\BNUM}{\begin{enumerate}}
\newcommand{\ENUM}{\end{enumerate}}
\newcommand{\BEL}[1]{\begin{equation}\label{#1}}
\newcommand{\EEL}{\end{equation}}

\newcommand{\state}{\mathbf{s}}

\newcommand{\action}{\mathbf{a}}

\newcommand{\statespace}{\mathcal{S}}
\newcommand{\actionspace}{\mathcal{A}}

\newcommand{\policy}{\pi}

\newcommand{\nens}{{N_\text{ensemble}}}

\newcommand{\BA}{\begin{array}}
\newcommand{\EA}{\end{array}}

\DeclareMathOperator*{\expec}{\mathbb{E}}

%% file: sections/intro.tex
\section{Introduction}
\label{sec:intro}
Agile, robust, and capable robotic skills require careful controller design and validation to work reliably in the real world. Reinforcement learning offers a promising alternative, acquiring effective control strategies directly through interaction with the real system, potentially right in the environment in which the robot will be situated. Although large-scale robotic reinforcement learning experiments in the real world have been described in a number of prior works~\cite{Kalashnikov2018QTOptSD, Levine2018LearningHC, Dasari2019RoboNetLM, Kalashnikov2021MTOptCM, Ebert2022BridgeDB}, many other researchers have sought to sidestep the need for real-world training over concerns about sample efficiency, often going to great lengths to design sophisticated simulation systems~\cite{Cutler2014ReinforcementLW, Sadeghi2017CAD2RLRS, Rajeswaran2017EPOptLR, Tobin2017DomainRF, Peng2018SimtoRealTO} and transfer learning methods~\cite{Yu2019SimtoRealTF, Hwangbo2019LearningAA, Xie2019LearningLS, Yu2020LearningFA, Peng2020LearningAR, Kumar2021RMA, Wang2017LearningTR, Finn2017ModelAgnosticMF, Rakelly2019EfficientOM, Song2020RapidlyAL}. For example, learning to solve a Rubik's cube with a robotic hand required 13 thousand years' worth of \emph{experience}~\cite{OpenAI2019SolvingRC}, which amounted to several months of wall-clock time using distributed training in simulation. In robotic locomotion, Rudin et al.~\cite{Rudin2021LearningTW} also utilize distributed training to collect 80 hours' worth of simulated experience to train an ANYmal robot to walk in 20 minutes. Using the A1 quadrupedal robot as we do, Kumar et al.~\cite{Kumar2021RMA} use $1.2\cdot10^9$ samples to train a robust controller to enable locomotion across many real-world terrains. This corresponds to roughly 4.5 months' worth of cumulative experience---without accounting for the often significant overhead required to train in the real world---that can be acquired in 1 day using an appropriate simulator. In this paper, we focus specifically on the task of robotic locomotion and ask: just how efficiently can we implement fully model-free deep RL algorithms?
\begin{table*}[ht]
\centering
\resizebox{\textwidth}{!}{
\scriptsize
\begin{tabular}{lcccccccc}
\toprule
                     &            \multicolumn{4}{c}{\textbf{Experimental Design}}         & \multicolumn{4}{c}{\textbf{Training Statistics}} \\
                                     \cmidrule(lr){2-5}                                      \cmidrule(lr){6-9}
                     &             &    &  & & \multicolumn{2}{c}{\textbf{Simulation}} & \multicolumn{2}{c}{\textbf{Real World}} \\
                                                       \cmidrule(lr){6-7}                           \cmidrule(lr){8-9}
\textbf{}      & Hardware & Actions & Resets & Terrains & Samples & Hours & Samples & Hours  \\
\midrule
Ours     & A1  & PD targets & Learned & In/Outdoor & 0 & 0 & $20\cdot10^3$ & $0.\overline{3}$ \\
Wu et al.~\cite{Wu2022DayDreamerWM}  & A1 & PD targets & None & Indoor & 0  & 0       & $72\cdot10^3$ & 1  \\
Smith et al.~\cite{Smith2022Finetune}  & A1 & PD targets & Learned & In/Outdoor & $10^6$ & N/A & $22.5\cdot10^3$ & 1\\
Kumar et al.~\cite{Kumar2021RMA}  & A1 & PD targets & N/A & In/Outdoor & $1.2\cdot10^9$ & 24 & 0 & 0 \\
Rudin et al.~\cite{Rudin2021LearningTW} & ANYmal  & PD targets & N/A &  Indoor & $14.7\cdot10^6$ & $0.\overline{3}$ & 0 & 0 \\
Lee et al.~\cite{Lee2020LearningQL}    & ANYmal & PMTG~\cite{Iscen2018PoliciesMT} parameters & N/A & In/Outdoor & N/A & 16 & 0 & 0 \\
Ha et al.~\cite{Ha2020LearningTW}       & Minitaur & PD targets & Engineered & Indoor & 0 & 0 & $60\cdot10^3$ & 1.5\\
Haarnoja et al.~\cite{Haarnoja2019LearningTW}       & Minitaur & PD targets & Manual & Indoor & 0 & 0 & $160\cdot10^3$ & 2 \\
Yang et al.~\cite{Yang2019DataER}       & Minitaur & PMTG~\cite{Iscen2018PoliciesMT} parameters & Unknown & Indoor & 0 & 0 & $45\cdot10^3$ & $0.1\overline{6}$ \\
\bottomrule
\end{tabular}%
}
\caption{Overview of experimental details (hardware platform used, the kinds of actions, and access to resets) and data requirements of the works most relevant to ours (ordered chronologically). We list the approximate numbers reported for the tasks that most closely resemble ours (walking forward). For ``Training Statistics" we list the data used for training, i.e., we do not include data used in evaluation (that may be used in few-shot adaptation~\cite{Kumar2021RMA}), and the wall-clock time associated with collecting that data either in simulation or in the real world. Some earlier works demonstrated learning in the real world on the Minitaur~\cite{Kenneally2016DesignPF}, a relatively simple legged robot with neither shoulder nor hip abduction, in controlled, lab settings. RL with more complex robots~\cite{Hutter2016ANYmalA} often used large amounts of simulation data. More similar in physical capabilities to the ANYmal and in accessibility to the Minitaur, the A1 robot has also been used to study real-world deployment in recent works.}
\label{tab:cmprs}%
\vspace{-.5cm}
\end{table*}%
Perhaps surprisingly, we find that with several careful design decisions in terms of the task setup and algorithm implementation, it is possible for a quadrupedal robot to learn to walk from scratch with deep RL in under 20 minutes, across a range of different environments and surface types. Crucially, this does not require novel algorithmic components or any other unexpected innovation, but rather careful implementation of one of several existing algorithmic frameworks (and indeed multiple algorithms can work well), combined with modern optimized deep learning packages and a number of careful design decisions for the MDP formulation of the locomotion task. This result runs counter to the principles articulated in several prior works, which suggest either than simulated training is critical for robotic locomotion because the training times are too long~\cite{Tan2016SimulationbasedDO, Liu2018LearningBD, Lee2019ScalableMH, Peng2018DeepMimicED, Rudin2021LearningTW}, that demonstration data is needed to overcome local optima challenges~\cite{Peng2018DeepMimicED, Peng2020LearningAR}, or that more sophisticated algorithmic frameworks with model-based RL~\cite{Yang2019DataER, Wu2022DayDreamerWM} or hand-designed movement primitives~\cite{Kohl2004PolicyGR, Tedrake2004StochasticPG, Endo2005LearningCS, Choi2019TrajectorybasedPP, Yang2019DataER} are necessary for real-world training.

Our main contribution is an empirical demonstration that current deep RL methods can effectively learn quadrupedal locomotion directly in the real world in under 20 minutes. While our results largely build on existing methods, we demonstrate for the first time that a careful combination of existing components can enable direct real-world learning of locomotion skills with drastically lower training periods than reported in prior work. Our evaluation includes real-world training in four different locations (one indoor, three outdoor), and detailed simulated analysis to understand the relative importance of different design choices.

%% file: sections/relatedwork.tex
\vspace{-.15cm} 
\section{Related Work}
\label{sec:relatedwork}
Roboticists have traditionally designed controllers for quadrupedal locomotion using a combination of footstep planning, trajectory optimization, and model-predictive control (MPC) \cite{Hutter2016ANYmalA, Park2017BoundingCheetah, Bledt2018MITC3, Katz2019MiniCA}. Learning provides an appealing alternative to classic model-based methods, as it avoids the need for intricate modeling and extensive domain knowledge about the locomotion task, instead allowing the learning algorithm to discover a gait that works well for a given robot. Furthermore, learning in the real world allows the robot to improve over time with ``on-the-job'' experience.

A major question in this line of work is whether RL methods could ever be efficient enough to learn dynamic locomotion in diverse real-world environments with more complex robots. Indeed, some works have sought to sidestep this challenge by using simulation to learn controllers that are directly transferred to the real world~\cite{Tan2016SimulationbasedDO, Chebotar2019ClosingTS, Du2021AutoTunedST, Tan2018SimtoRealLA, Xie2019LearningLS, Peng2018SimtoRealTO, sadeghi2016cad2rl, Lee2020LearningQL, Yang2021FastAE, Rudin2021LearningTW} or learning policies that can use small amounts of data when deployed in the real world to search among the space of policies it has for a suitable one~\cite{He2018ZeroShotSC, Yu2019SimtoRealTF, Hwangbo2019LearningAA, Xie2019LearningLS, Yu2020LearningFA, Peng2020LearningAR, Kumar2021RMA}. Still, these methods rely entirely on the strategies the robot was able to learn while training in simulation. 
So, the training conditions must be designed and implemented such that the learning under them captures the behaviors that are transferable to anticipated test-time conditions in the real-world. While this approach to learning controllers is often sufficient, it is non-trivial to foresee all the possible situations the robot may encounter when deployed in the real world and engineer the appropriate training environments in simulation to prepare for them. Furthermore, these learned models are fundamentally limited to those training experiences---they cannot perfectly generalize when they are tested in situations that differ enough from their training experience, but these unexpected situations are likely to arise in the real world that is highly complex, unstructured, and unpredictable.

Early work in instead learning \emph{directly} in the real world explored utilizing higher level action spaces~\cite{Kohl2004PolicyGR, Tedrake2004StochasticPG, Endo2005LearningCS, Luck2017FromTL, Choi2019TrajectorybasedPP, Yang2019DataER, Yang2022SafeRL} or fine-tuning~\cite{Smith2022Finetune} to do real-world training. Perhaps surprisingly, we show that careful implementation of already known model-free RL techniques can enable learning to walk from scratch using low-level control, such as PD targets, in several real-world environments. Next, we detail prior work in using model-free RL for real-world training of locomotion policies, followed by model-based work, and finally, systems aspects relevant to contextualizing these approaches. For a tabular overview comparing the assumptions, requirements, and results of the works most relevant to ours, see Table~\ref{tab:cmprs}.

% other model-free methods
\paragraph{Model-free learning} Much of the early work on learning from scratch in the real world applied model-free policy gradient methods to optimize parameters of pre-defined motions on relatively simple hardware. Kohl et al.~\cite{Kohl2004PolicyGR} learn parameters of a pre-defined open-loop trajectory generator to control a Sony AIBO quadruped, and Tedrake et al.~\cite{Tedrake2005LearningTW} learn a feedback controller to improve the control of a passive dynamic walker whose design is such that it walks when control is disabled, using policy gradient methods. Luck et al.~\cite{Luck2017FromTL} achieved sample-efficient learning by leveraging periodicity to similarly learn few parameters to modulate a pre-defined controller of a finned robot. Choi et al.~\cite{Choi2019TrajectorybasedPP} also use a stochastic policy gradient method learn to control a Snapbot quadrupedal robot by learning a distribution over Gaussian random paths that define fixed length joint trajectories that are then realized by an open-loop controller. Yang et al.~\cite{Yang2022SafeRL} learns a model-free RL policy in the real world which outputs gait parameters and foot placements that are realized by a low-level MPC controller. While very effective for learning walking motions in the real world by shaping exploration and ensuring safety, using high level action spaces limits the types of skills that can be learned. Recent works~\cite{Haarnoja2019LearningTW, Ha2020LearningTW} have utilized off-policy methods~\cite{Haarnoja2018SoftAO} to perform sample-efficient learning on a Minitaur quadrupedal robot (2 active DOF per leg) by directly outputting target joint positions. Training requires roughly 2 hours (160k control steps) to learn to walk forward and 1.5 hours (60k control steps per direction) to learn to walk forward and backward, respectively. Our work also uses off-policy model-free RL to learn from scratch in the real world---uniquely, though, we train an A1 quadrupedal robot (3 active DOF per leg) not only in lab settings, but also on outdoor irregular terrains, in just 20 minutes. 

% model-based methods
\paragraph{Model-based learning} There have also been a number of model-based RL methods applied to enable an A1 robot to learn how to walk on flat, solid ground~\cite{Yang2019DataER, Wu2022DayDreamerWM}. Yang et al.~\cite{Yang2019DataER} use trajectory generators~\cite{Iscen2018PoliciesMT}, which output smooth, periodic leg trajectories and use a model-based method to learn to modulate these trajectories. Concurrent work by Wu et al.~\cite{Wu2022DayDreamerWM} uses a latent dynamics model to generate additional training data in order to reduce the burden on collecting real-world samples~\cite{Hafner2020DreamTC} in order to learn to walk with low-level PD targets as actions within an hour. In contrast, we use a learned policy to automatically provide resets, and our robot learns to walk within 20 minutes in five environments, three of which are natural outdoor terrains, using a simple model-free method.

% system aspects (e.g., async training) relatively briefly.
\paragraph{Systems considerations} RL is known to require lots of data to learn complex behaviors. For example, a state-of-the-art blind quadrupedal locomotion policy trained completely in simulation required 12 hours~\cite{Lee2020LearningQL}; however, real-world data is significantly more expensive to collect. One line of work in making RL more efficient has been to eliminate the data collection bottleneck, using upwards of thousands of workers to collect experience simultaneously and consolidating the information for policy updates~\cite{Nair2015MassivelyPM,Babaeizadeh2017ReinforcementLT, Espeholt2018IMPALASD}. In the locomotion domain, Rudin et al.~\cite{Rudin2021LearningTW} use this parallelism to train a simulated ANYmal quadruped to walk on uneven terrain---in 20 minutes of wall-clock time---and then deploy the learned policy in the real world. However, massively parallel data collection is not often feasible in the real world, so in order to there has been a focus on enabling sample-efficient methods with asynchronous pipelines. Due to computational costs many prior works, in fact, claim asynchronous training to be necessary for real-world training~\cite{Gu2017DeepRL, Haarnoja2018SoftAO, Haarnoja2019LearningTW, Zhang2019AsynchronousMF, Wu2022DayDreamerWM}. Most similar in spirit to our work, Haarnoja et al.~\cite{Haarnoja2019LearningTW} have a three-part asynchronous training pipeline, with jobs dedicated to data collection, motion capture for state and reward estimation. Follow-up work~\cite{Ha2020LearningTW} achieved superior sample efficiency (approximately half the samples required) with the same underlying algorithm by performing synchronous training at a per-step basis (as opposed to episodic, asynchronous training), thereby reducing overall wall-clock time by a similar factor. Recently, more efficient off-policy model-free methods have been enabled by allowing more gradient steps to be taken per sample collected, placing a larger burden on the computation. In our work, we find that we are able to support synchronous, per-step training in our implementation (see Subsection \ref{sec:method}), achieving learning in 20 minutes on a real robot using a single GPU laptop.

%% file: sections/preliminaries.tex
\section{Fast and Simple RL for Real-World Robots}
\label{sec:preliminaries}

In this section, we will describe the algorithmic framework that we use to enable robots to learn to walk in the real world. Our algorithmic framework is not novel, and is based on standard Q-function actor-critic methods~\cite{sutton2018reinforcement}, building most directly on DroQ~\cite{Hiraoka2021DropoutQF}, which extends the SAC algorithm~\cite{Haarnoja2018SoftAO} with Dropout~\cite{Srivastava2014DropoutAS} and Layer Normalization~\cite{ba2016layer}. However, for the task considered in this work, we only see significant improvement from adding Layer Normalization and increasing the update to data ratio (Section~\ref{sec:sim-experiments}).
We emphasize that our result is not enabled so much by any one algorithmic component (though the algorithm is also important!), but rather careful implementation and task setup, which we discuss in Section~\ref{sec:system}.

\subsection{Preliminaries} Learning to walk can be formalized as an infinite horizon Markov Decision Process (MDP), which is defined by a tuple $(\statespace, \actionspace, p_0, p, r, \gamma)$ where $\statespace \subset \mathbb{R}^n$ is the state space, $\actionspace \subset \mathbb{R}^m$ is the action space, $p_0(\cdot)$ is the initial state distribution, $p(\cdot|s, a)$ is the transition function, $r: \statespace \times \actionspace \rightarrow \mathbb{R}$ is the reward function, and $\gamma \in [0, 1)$ is the discount factor. The goal of RL is to optimize the expected discounted cumulative return induced by the policy $\pi: \statespace \rightarrow \actionspace$:
$$
\expec\left[\sum_{t=0}^\infty \gamma^t r(s_t, a_t)  | s_0 \sim p_0(\cdot), a_t \sim \pi(\cdot| s_t), s_{t+1} \sim p(\cdot| s_t, a_t)\right].
$$

We consider actor-critic methods consisting of interleaved policy evaluation and improvement steps. During policy improvement, we fit a critic to estimate the discounted returns of the current training policy starting at state $s$ and action $a$:
\begin{align*}
J(\theta) &= \expec_{(s, a, s')\sim D} [(Q_\theta(s,a) - y(s,a,s'))^2] \\ 
y(s,a,s') &= r(s,a) + \gamma Q_{\theta'}(s', a') \mbox{ where } a' \sim \pi_\eta(\cdot|s') 
\end{align*}
where $Q_{\theta'}$ is a target network with weights updated via exponential moving average. Then, the critic is used to improve the policy by maximizing the objective:
\begin{align*}
J(\eta) = \expec_{\substack{s \sim D \\ a \sim \pi_\eta(\cdot|s)}}[Q_\theta(s, a)].
\end{align*}
The most widely-used off-policy RL algorithms for continuous control, such as SAC and DDPG \cite{Haarnoja2018SoftAO, Xu2020DeepDP}, are commonly trained by making one critic update and one policy update after each environment step. Since the update time can exceed what is allowed by the robot's control frequency, sometimes real-world training is performed by collecting an entire trajectory and then updating the critic and policy for a number of steps equal to the length of the trajectory~\cite{Haarnoja2019LearningTW}.
Next we will discuss the simple changes to these common practices we make that has enabled our result.

\subsection{Efficient Model-Free RL}
\label{sec:method}

Actor-critic methods have recently become significantly more sample-efficient by improving the training of the critic, thereby allowing more updates to the critic network for the same amount of training data. For example, REDQ~\cite{Chen2021RandomizedED} extends SAC~\cite{Haarnoja2018SoftAO} by utilizing a large ensemble of critics and computing target values by minimizing over a random subset of them. DroQ \cite{Hiraoka2021DropoutQF} similarly allows for a higher update to data ratio by regularizing the critic networks with dropout~\cite{Srivastava2014DropoutAS} and layer normalization~\cite{Ba2016LayerN}.  
What all of these works have in common is that they add some sort of regularization or normalization method (or both) to mitigate the tendency that off-policy bootstrapped critic updates have to produce overestimation and to overfit to the current target network~\cite{Chen2021RandomizedED, Hiraoka2021DropoutQF}.
These techniques allow the algorithm to take significantly more gradient steps on the critic after each environment step, which in turn leads to significantly more sample-efficient learning. As we will discuss in our analysis in Section~\ref{sec:sim-experiments}, we find that a variety of regularization and normalization approaches all lead to the critical jump in improvement over the baseline method. These results suggest that the key is not any one specific critical choice, but the general principle of augmenting actor-critic RL with regularization or normalization. These algorithms use up to 20 times the number of critic updates to speed up learning with respect to the number of samples collected, but this increases their computational cost such that it actually supersedes data collection as the primary bottleneck in real-world training.

Because of this, a na\"{i}ve implementation cannot train as fast as the samples are collected. Prior work has addressed this either by performing asynchronous training~\cite{Haarnoja2019LearningTW, Wu2022DayDreamerWM} or training in-between trials~\cite{Smith2022Finetune}. Both options add a delay between the agent interacting with the environment and learning from the samples, which slows down training. Our choice of algorithm and implementation is aimed at enabling real-time synchronous training, which we expand on in Section~\ref{sec:sim-experiments}. For implementation, we use JAX~\cite{jax2018github}, a modern machine learning framework that performs just-in-time compilation to optimize execution significantly (see Section~\ref{sec:real-experiments} for a discussion of the practical aspects).

%% file: sections/system.tex
\section{System Design}
\label{sec:system}
We design our system so as to prioritize performing fast training in unstructured real-world environments. We use a relatively low-level action space---rather than using pre-defined motion primitives as discussed is used in many prior works (see Section~\ref{sec:relatedwork})---and use only proprioceptive information---so as to be able to train anywhere, without an instrumented motion capture system. Of course, the MDP definition and implementation often has a large impact on learning, e.g., we found the design of the action space to be particularly important. In the remainder of this section, we detail these design decisions.

For our robot platform, we use the A1 robot from Unitree and build our simulation using MuJoCo~\cite{todorov2012mujoco} and DM Control~\cite{tunyasuvunakool2020}. The policy $\policy_i$ and Q-function $\{Q_{\theta_i}\}_{i=1}^\nens$ are modeled using separate fully-connected neural networks that are constructed and trained using JAX~\cite{jax2018github}. 

\subsection{State and Action Spaces} 
For learning locomotion controllers, the robot's position is used in order to provide reward supervision~\cite{tunyasuvunakool2020}, and privileged information is often used to train policies in simulation~\cite{Lee2020LearningQL, Kumar2021RMA}. As such, such policies cannot trivially be further trained in the real world. When training in the real world, motion capture systems have been used to localize the robot~\cite{Haarnoja2019LearningTW, Ha2020LearningTW}. However, in order to train in the wild, the robot must be able to receive feedback purely from its onboard sensors. In terms of actions, for generality, we parameterize the policy to directly output joint targets rather than rely on pre-defined gaits or trajectory generators.

The state $\state_t$ contains the root orientation, root angular velocity, root linear velocity, joint angles, joint velocities, binary foot contacts, and the previous action. For the root orientation, we include the roll and pitch. For angular velocity, we include roll, pitch, and yaw information (as to penalize excessive turning). We use an onboard linear velocity estimator which combines integrated acceleration and leg velocity estimated with forward kinematics via Kalman filter that was shown to suffice for reward supervision in outdoor environments~\cite{Smith2022Finetune}. 
Actions $\action_t$ are PD position targets for each of the 12 joints and applied at a frequency of 20Hz. We define the action space for every leg as $[p - o, p + o]$ where $p$ corresponds to default motor angles and $o$ is an action offset. Following Fu et al.~\cite{Fu2021MinimizingEC}, we use $o=[0.2, 0.4, 0.4]$. We confirm that constraining the action space is crucial for training the agent with reinforcement learning in Section~\ref{sec:sim-experiments}. Moreover, we found that the default motor angles, which correspond to the initial robot pose and, therefore, define the action space, also impact exploration significantly. In the simulator, we used $p=[0.05, 0.7, -1.4]$; however, during the early experiments in the real world, we found that $p=[0.05, 0.9, -1.8]$ promotes safer exploration on a real robot since this configuration leads to fewer failures.
We use a position controller where the torques commanded to the robot are computed as $\tau = K_p(q^*-q) - K_d\cdot \dot{q}$.
In this case, $q^*$ and $q$ define the desired and current positions correspondingly, while $v$ defined motor velocities. $K_p$ and $K_d$ define motor gains and damping.

\subsection{Reward Function}
Prior works have reported using very complex reward functions consisting of upwards of tens of terms to give rise to a desired motion~\cite{Lee2019RobustRC, Lee2020LearningQL, Peng2020LearningAR, Fu2021MinimizingEC}. We found that with the state and action space choices described above, even a simple reward function was sufficient to produce naturalistic gaits. We define the reward function following standard DM Control~\cite{tunyasuvunakool2020} locomotion tasks, where the agent is provided with a constant reward within a target velocity interval, outside of which the reward linearly decays to 0.

$$
r(s,a) = r_v(s,a) - 0.1 v_{yaw}^2
$$

where $v_{yaw}$ is an angular yaw velocity and

\[
r_v(s, a) = 
\begin{cases}
    1, & \text{for } v_x \in [v_{t}, 2 v_{t}] \\
    0, & \text{for } v_x \in (-\infty, -v_{t}] \cup [4 v_{t}, \infty) \\
    1-\frac{|v_x - v_{t}|}{2v_{t}}, & \text{otherwise.}
\end{cases}
\]
where $v_{t}$ is the target velocity, while $v_x$ is a forward linear velocity in the robot frame. Our choice of the reward function is motivated by simplicity.

During early experiments with the real robot, we found that using the forward velocity in the robot's local frame caused it to dive forward as falling quickly onto its head does correspond to high forward velocity from the robot's perspective. However, our goal is for the robot to move forward in the global frame, parallel to the ground plane, so we instead project the forward velocity onto the ground plane.

We relied on rewards estimated via an HTC Vive VR system for the initial indoor experiments but not viable for outdoor experiments. As such, we switched to a Kalman filter-based velocity estimator. We compared the velocity provided by this built-in estimator to estimates from a VR system and found that the estimator is significantly less accurate and systematically underestimates the velocity. While this causes a stark difference in the rewards attained in simulated and real-world experiments, we nonetheless found it quite feasible to learn effective gaits in the real world.

We made further adjustment to accommodate moving the robot to keep it from leaving the training area with minimal interruption, we simply continue training while providing a reward of 0 when we detect that someone has lifted the robot to move it (i.e. when there are no foot contacts detected). Otherwise, the robot trains continuously, terminating only when the robot's roll or pitch exceeds 30 degrees. To reset the robot in the real world, we use the open-source reset policy from Smith et al.~\cite{Smith2022Finetune}.

%% file: sections/experiments.tex
\section{Simulation Analysis}
\label{sec:sim-experiments}
This section presents simulated comparisons of design decisions and SAC variants we considered in this work. Since our goal is to run training on a real robot, we aim for design decisions and algorithms that lead to improved stability and sample efficiency. 
To match the real-world setup, we simulate the official A1 model in MuJoCo, and used the same position controller and rewards as discussed in \Cref{sec:method}. We provide the exact model we used on the project website.

\begin{figure}[t]
\centering
  \hspace{-.3cm}
  \begin{minipage}[b]{.25\textwidth}
    \subcaption[]{Damping}\label{fig:sim-results_a}  
    \vspace{-.2cm}
    \includegraphics[align=c,height=1.1in]{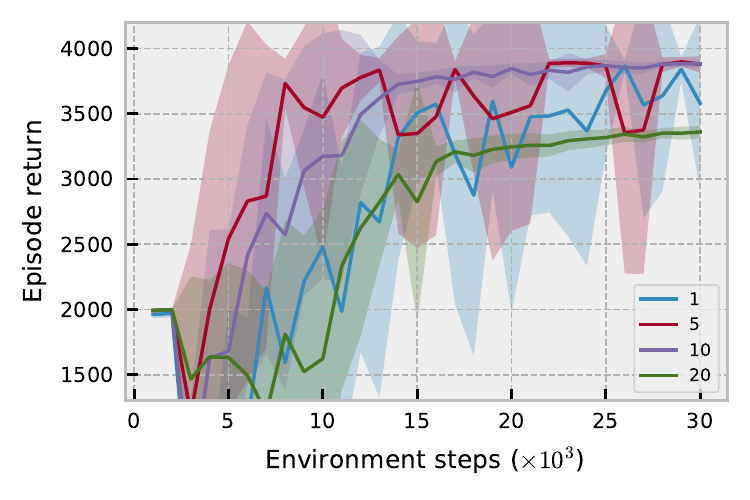}\hspace{-.25cm}
  \end{minipage}\hspace{-.25cm}
  \begin{minipage}[b]{.25\textwidth}
    \subcaption[]{Setup ablations}\label{fig:sim-results_b}
    \vspace{-.2cm}
    \includegraphics[align=c, height=1.1in]{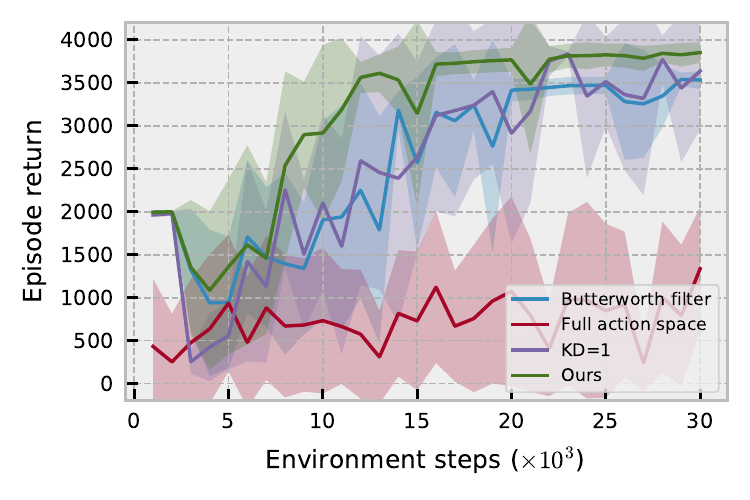}
  \end{minipage}\\
  \hspace{-.3cm}
  \begin{minipage}[b]{.25\textwidth}
   \subcaption[]{SAC variants}\label{fig:sim-results_d}
   \vspace{-.2cm}
   \includegraphics[align=c, height=1.1in]{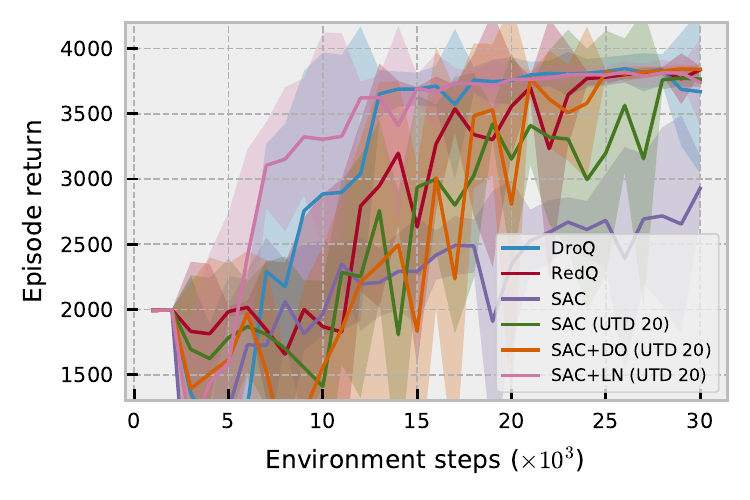}
  \end{minipage}\hspace{-.25cm}
    \begin{minipage}[b]{.25\textwidth}
  \subcaption[]{Synchronous training}\label{fig:sim-results_c}
  \vspace{-.2cm}
    \includegraphics[align=c, height=1.1in]{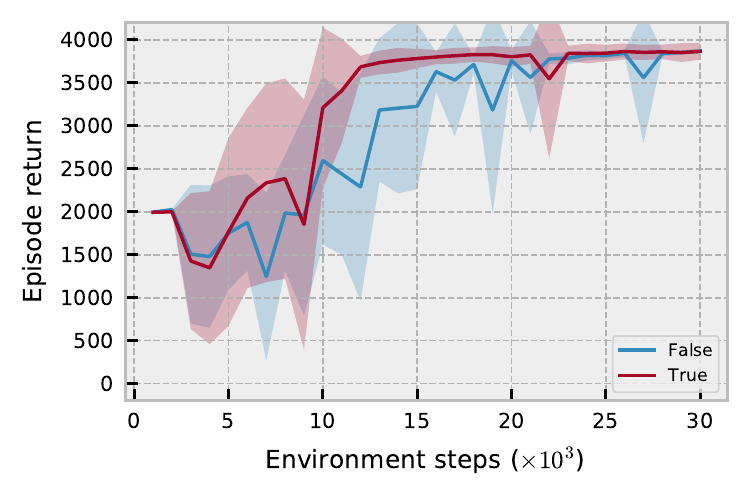}\hspace{-.1cm}
  \end{minipage}
\caption{\footnotesize Experimental evaluation of
(a) performance for different value of the damping parameter for the position PD controller;
(b) ablations of various task setup choices;
(d) regularization and normalization methods for efficient RL;
(c) the effect of the frequency of policy updates (between time-steps versus episodes). Each curve and shaded region represents the average and standard deviation, respectively, across 10 random seeds.
}
\label{fig:sim-results}
\end{figure}

\paragraph{MDP formulation} First, we observe that the value of damping for the position controller used for the robot significantly impacts learning (see \Cref{fig:sim-results_a}). Small damping values ($Kd=1$) lead to instabilities, which are not desirable for a policy executed on a real robot, while large values prevent the agent from reaching the target velocity ($Kd=20$). Therefore, for the remaining ablations, we used the value of damping set to $10$. We also evaluate other design decisions in \Cref{fig:sim-results_b}. In particular, we confirm the efficacy of constraining the action space: we observe that the simulated agent cannot make any progress in the unconstrained action space, while constraining the space leads to stable training and does not prevent the agent from reaching the target velocity. Finally, we note that applying a low-pass filter to the PD targets to promote smoothness, as is common practice~\cite{Ha2020LearningTW, Peng2020LearningAR, Smith2022Finetune}, degrades learning efficiency. This is perhaps unsurprising, as the filter creates a dependency on action history that violates the Markovian assumption.

\begin{figure*}[t]
\centering
\includegraphics[width=\linewidth]{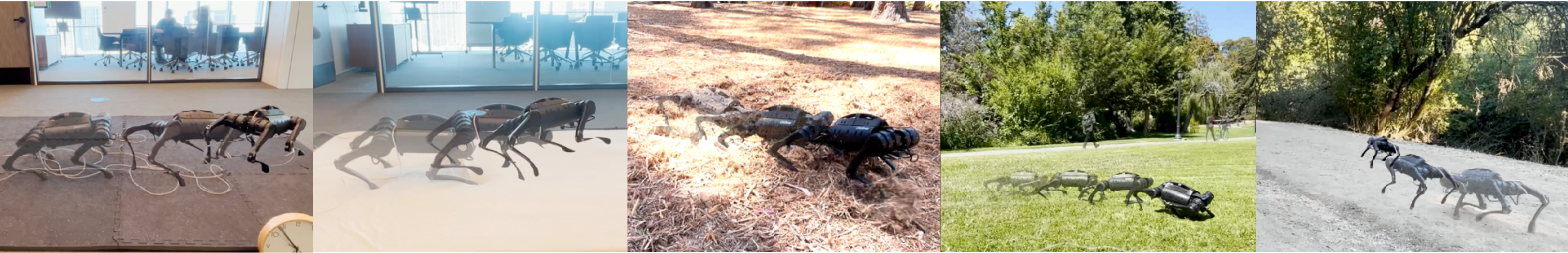}
\captionof{figure}{Examples of learned gaits acquired on a variety of real-world terrains. Left to right:  flat, solid ground covered in dense foam mats; a 5cm memory foam mattress; loose ground comprised of eucalyptus bark; a grassy lawn; a gently sloped hiking trail.}
\label{fig:locomotion-envs}
\end{figure*}

\paragraph{Algorithms}
Our goal is to train a robot to walk in the real world as efficiently as possible, and efficiency includes computational complexity, sample complexity, and total wall-clock time. As discussed in Section~\ref{sec:method}, we utilize off-policy model-free actor-critic methods, and investigate a variety of regularization and normalization techniques to accelerate them. All of the methods we consider are based on SAC~\cite{Haarnoja2018SoftAO}. We present an extensive comparison in Figure~\ref{fig:sim-results_c}, with the aim of understanding which ingredients are important for attaining the requisite sample efficiency for training in the real world. Standard SAC (purple) with an update-to-data (UTD) ratio of 1 takes one gradient step on the critic for every one time step of data collection. Efficiency can be increased by taking more gradient steps, and we show standard SAC with a larger UTD ratio of 20 (dark blue) as well. We see that na\"{i}vely increasing the number of critic updates made per time-step improves sample efficiency, but still requires roughly 30k samples, which would amount to roughly 30 minutes' worth of data, to reach the target velocity.
We implement REDQ~\cite{Chen2021RandomizedED} by modifying SAC with UTD ratio of 20 to use random subsets of a larger ensemble of networks in order to calculate target values, and we see this regularization indeed leads to improvement (roughly 5k fewer time-steps, or 5 minutes wall-clock time, required). As noted in DroQ~\cite{Hiraoka2021DropoutQF}, REDQ is more computationally expensive due to the large ensemble, and regularizing the critic with a combination of layer normalization and dropout can lead to similar benefits at lower compute cost (blue). However, we can also consider layer norm (pink) and dropout (orange) in isolation (each also with a UTD ratio of 20). Perhaps surprisingly, dropout alone already performs similarly to REDQ, whereas layer norm (even without dropout) leads to even better performance. From these results, we can conclude that a variety of regularization or normalization methods, if implemented and applied carefully, can all achieve a similar level of improvement in performance over their underlying algorithm \emph{in our setup}. That is, the important thing is not any single specific regularization technique, but the use of any suitable regularization so as to enable SAC to effectively use higher UTD ratios.

We also compare updating the agent between episodes and after every environment step and notice that getting immediate feedback leads to more stable training and faster convergence (see \Cref{fig:sim-results_d}). In order to facilitate this kind of training synchronously, the updates must be inexpensive enough to be able to perform them between time-steps (of which there are 20 per second). As such, we favor using the less computationally expensive DroQ variants over others in the real world.

\section{Learning in the Real World}
\label{sec:real-experiments}
We aim to evaluate the efficacy of our approach by answering the following through our experiments:

\begin{enumerate}[(1)]
    \item How quickly and consistently can the robot learn to walk in the real world using model-free RL?
    \item Can this approach enable a robot to learn to walk not only on flat ground in the lab, but also `in the wild'?
\end{enumerate}
In order to understand the variance across random seeds and small changes in real-world conditions (e.g., the status of the hardware, how many times the robot was redirected, etc.), for one of our experiments, we train the robot four times in the same controlled environment. 
To study (2), we train the robot in four additional terrains, three of which are outdoors. 
Lastly, we discuss our findings from using the design decisions, tuned in simulation as described in \autoref{sec:system}, to train in the real world.
Videos of all real-world training along with code to reproduce our results can be found on the project website\footnote{\tt \url{https://sites.google.com/berkeley.edu/walk-in-the-park}}.

\subsection{Setup}
\label{sec:setup}
We conduct experiments with five terrains (see \autoref{fig:locomotion-envs}), each of which possesses unique characteristics:

\begin{enumerate}[(1)]
    \item \textit{Flat, Solid Ground:} We placed tiles of dense foam on the floor for protection that make for a high-friction surface.  
    \item \textit{Memory Foam Mattress:} The other indoor terrain we test on is a $5$cm-thick memory foam mattress. The robot's feet sink fairly deep into the surface---see, e.g., the depression in the surface of the mattress made by the robot in the earliest frame in~\autoref{fig:locomotion-envs} (second from left), making walking difficult and requiring a unique gait to attain ground clearance. 
    \item \textit{Mulch:} A layer of bark (about a foot deep) makes walking especially difficult as the terrain is not only heavily irregular but also obstructive. As shown in~\autoref{fig:locomotion-envs} (middle), the robot naturally sinks such that the lower half of its legs are submerged in the bark. Thus, it must learn to churn through it in order to move.
    \item \textit{Lawn:} The lawn presents a lower friction, cushioned, deformable walking surface.
    \item \textit{Hiking Trail:} Here the surface material is a compact, dry dirt. As shown in~\autoref{fig:locomotion-envs} (right), the trail is at a slight incline and irregular (tree roots, pebbles, troughs, etc.), presenting additional challenges.
\end{enumerate}

To be able to train outdoors, we use a laptop (Origin EON15-X) for training equipped with a single NVIDIA GeForce RTX 2070 GPU. For all experiments, we collect data for about 1 minute (1000 time-steps) by sampling from the action space uniformly at random. We then train synchronously, making a policy update (20 critic updates and an actor update) in between each action executed on the robot every 0.05 seconds. On a standard workstation equipped with a NVIDIA GeForce RTX 2070 GPU, our initial JAX implementation of DroQ was capable of performing 700 critic updates per second, roughly corresponding to 35 time-steps per second. However, this was insufficient for training the agent on a laptop, which reiterates the importance of implementation for real-world training. For our final implementation, we jit 20 critic updates corresponding to one environment step, resulting in 2400 critic updates per second which corresponds to running training at 120 Hz. In contrast, for a baseline comparison, our best PyTorch implementation was not fast enough for synchronous training, only permitting an effective control frequency of 7.5 Hz.

\subsection{Results}

\begin{figure}
\centering
  \hspace{-.5cm}
  \includegraphics[align=c, height=1.2in]{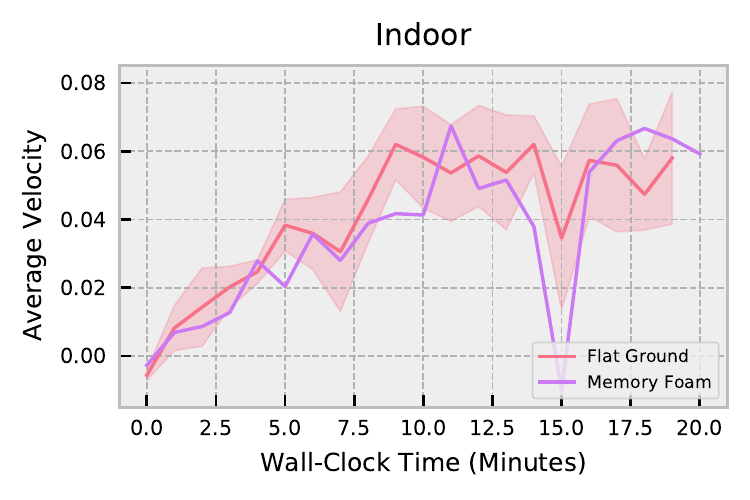}\hspace{-.2cm}
  \includegraphics[align=c, height=1.2in]{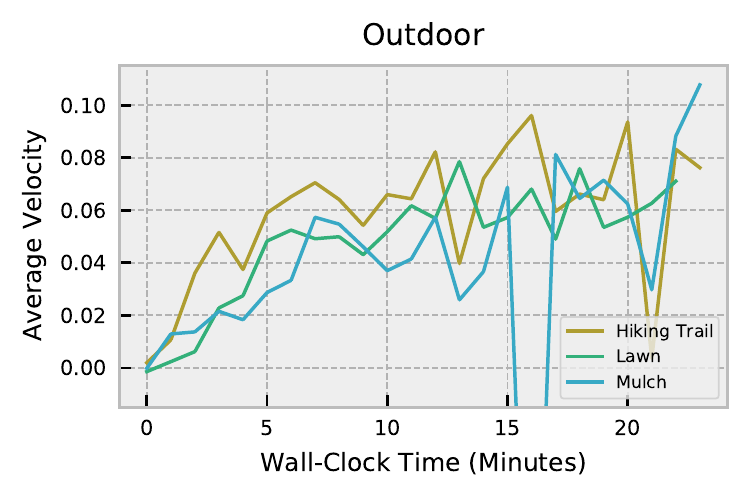} \hspace{-.2cm}
 \caption{\footnotesize Learning curves for all the real-world experiments showing the average velocity of the stochastic policy with respect to real-world, wall-clock time. Note that the robot runs continuously, training in between sending actions to the physical hardware. 1000 time-steps corresponds roughly to 1 minute of training time, all things considered.
 }
\label{fig:real-world-plots}
\end{figure}

We report the average velocity over intervals of 1000 time-steps (corresponding to 1 minute of wall-clock time) during training in \autoref{fig:real-world-plots}
with respect to the number of real-world samples collected during which the robot was on the ground (not including the times the robot was lifted and reoriented).
Per answering (1), we report the average and standard deviation (solid line and shaded region, respectively) across four runs on the flat, solid ground (pink). In all cases, the robot learns to walk in less than 20,000 samples (roughly 17 minutes' worth of data which amounts to 20 total minutes of wall-clock time due to various minor sources of overhead: jitting (about 1 minute), resetting and reorienting the robot, etc.). At the start of training, the robot mostly shuffles, slowly edging backwards. In 5 minutes, the robot learns to inch forward but is unstable. Within 10 minutes, the robot learns to take fairly large steps, but it has not learned to maintain balance while taking these larger steps. Finally, after 15 minutes of training, the robot adopts more conservative behavior in order to both walk and remain balanced.

Next, we train on the variety of challenging terrains detailed in Subsection~\ref{sec:setup}. On the memory foam, the robot makes its way out of the initial position it had sunk into during initial data collection (similar to the situation shown in the earliest frame of~\autoref{fig:locomotion-envs} (second from left)) within about 5 minutes. In contrast to the flat, solid ground, the robot falls much less frequently, but its feet often catch on the fabric cover and it needs to learn to walk in a way such that it avoids dragging the fabric with it in order to traverse it. On the thick layer of wood mulch, the initial data collection essentially digs the robot quite deep into the loose ground. In the first ten minutes, the robot learns to kick up its front feet in order to dig itself through, as seen most clearly in the latest frame of~\autoref{fig:locomotion-envs} (middle) where the robot is kicking its front, right leg up to pull itself forward. Furthermore, the terrain is irregular, yet it is able to quickly adapt its behavior. In all cases the robot learns to traverse each given terrain with roughly 20,000 samples (20 minutes).

%% file: sections/discussion.tex
\section{Conclusion}
\label{sec:conclusion}
We present our finding that we can train legged robots to walk via deep RL in real-world settings, such as grass, loose ground, forest trails, and mattresses, with about 20 minutes of training. We demonstrate that this can enabled using a high-quality implementation that is based on existing algorithmic ideas, by combining standard actor-critic algorithms with one of a number of different regularization strategies. We compare the different design choices in simulation and show that a variety of designs can work well, and then demonstrate in the real world that this leads to highly efficient and successful learning on a range of different terrains. Our empirical results show that real-world training of locomotion policies via RL can be significantly more practical than previously believed, and does not necessarily require significant deviations from existing practice in RL, but rather careful combination of current best practices. We hope that our work will serve to further encourage investigation of real-world RL in robotics.

%% file: sections/acknowledgements.tex
\subsection*{Acknowledgements}
This work was supported by the Office of Naval Research and DARPA RACER. Laura Smith is supported by NSF Graduate Research Fellowship. We thank Kevin Zakka and Vikash Kumar for their help with the A1 MuJoCo model and Philipp Wu for designing and printing a protective shell for the physical robot.